\documentclass[review,5p,twocolumn]{elsarticle}   

\usepackage{lineno,hyperref}
\usepackage{amssymb}
\usepackage{algorithm}
\usepackage{algorithmic}
\usepackage{amsmath}
\modulolinenumbers[5]

\begin{document}

\begin{frontmatter}

\title{Domain Adaptive Semantic Segmentation by Optimal Transport}
\tnotetext[mytitlenote1]{This research is supported by the National Natural Science Foundation of China (No. 11971296) and National Key R \& D Program of China (No. 2021YFA1003004)}
\tnotetext[mytitlenote2]{Fully documented templates are available in the elsarticle package on \href{http://www.ctan.org/tex-archive/macros/latex/contrib/elsarticle}{CTAN}.}


\author[mymainaddress]{Yaqian Guo}
\author[mymainaddress]{Xin Wang}
\author[address2]{Shaoyi Du}
\author[address3]{Ce Li}
\author[mymainaddress]{Shihui Ying\corref{mycorrespondingauthor}}
\ead{shying@shu.edu.cn}

\cortext[mycorrespondingauthor]{Corresponding author}

\address[mymainaddress]{Department of Mathematics, Shanghai University, Shanghai, 200444, China}
\address[address2]{Institute of Artificial Intelligence and Robotics, Xi’an Jiaotong University, Xi’an 710049, China}
\address[address3]{College of Electrical and Information Engineering, Lanzhou University of Technology, Lanzhou, 730050, China}

\begin{abstract}
Scene segmentation is widely used in the field of autonomous driving for environment perception, and semantic scene segmentation (3S) has received a great deal of attention due to the richness of the semantic information it contains. It aims to assign labels to pixels in an image, thus enabling automatic image labeling. Current approaches are mainly based on convolutional neural networks (CNN), but they rely on a large number of labels. Therefore, how to use a small size of labeled data to achieve semantic segmentation becomes more and more important. In this paper, we propose a domain adaptation (DA) framework based on optimal transport (OT) and attention mechanism to address this issue. Concretely, first we generate the output space via CNN due to its superiority of feature representation. Second, we utilize OT to achieve a more robust alignment of source and target domains in output space, where the OT plan defines a well attention mechanism to improve the adaptation of the model. In particular, with OT, the number of network parameters has been reduced and the network has been better interpretable. Third, to better describe the multi-scale property of features, we construct a multi-scale segmentation network to perform domain adaptation. Finally, in order to verify the performance of our proposed method, we conduct experimental comparison with three benchmark and four SOTA methods on three scene datasets, and the mean intersection-over-union (mIOU) has been significant improved, and visualization results under multiple domain adaptation scenarios also show that our proposed method has better performance than compared semantic segmentation methods.
\end{abstract}


\end{frontmatter}


\section{Introduction}
Scene segmentation is widely used in the field of autonomous driving for environment perception, as the scene contains a lot of complex semantic information, and 3S has received a great deal of attention. As one of core tasks in computer vision, it focuses on complete scene understanding of the surrounding environment in intelligent transportation systems. Unlike classical computer vision research, such as image classification, object detection and instance segmentation, current mainstream deep learning-based 3S uses algorithms to densely predict each pixel of an image and assign a label to each pixel. The results of deep semantic segmentation allow computers to have a more detailed and accurate understanding of images, and have a wide range of applications such as autonomous driving, medical imaging, and robot simulation \cite{8578878}. 

Traditional 3S methods directly conduct a mapping from the image space to the label space. They mainly include thresholding methods, K-means clustering, support vector machine (SVMs), edge-based approaches and region-growing methods. Besides, kernel method is adopted to deal with inseparable tasks through mapping current dimensional features to the high dimensional space. In recent years, as we known, the usage of deep learning algorithms for domain adaptation has received extensive attention in computer vision \cite{2016Deep,2017Label,2016Domain,2016Coupled}. Due to the excellent feature representation, CNN obtains remarkable results on image classification, segmentation, and detection. However, the divergence of natural scene images is large, which leads the transferring of labels is difficult. 

For this issue, DA offers an efficient way to achieve the labels transfer, which transfers the knowledge from the source domain to the target domain based on a small number of annotations. Unfortunately, due to the existence of the huge domain gap between the source and the target domains, we need transfer the knowledge learned on the labeled source domain to the unlabeled target domain, i.e., we should address the unsupervised domain adaptation (UDA) problem in 3S. 

\begin{figure}[!htbp]
\centering
\includegraphics[width=1.0\columnwidth]{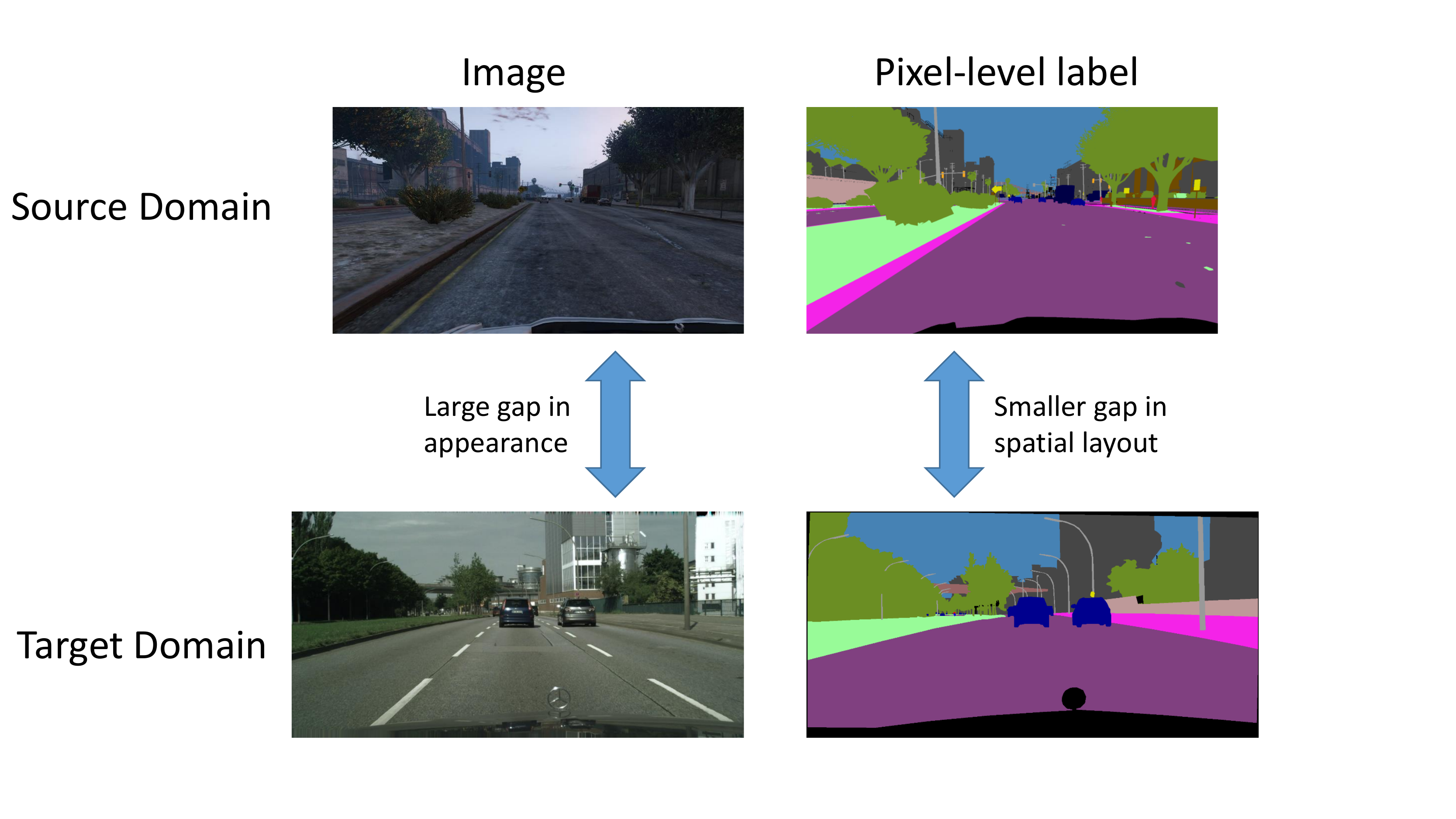} 
\caption{The motivation for distribution alignment in the output space. Even if the two images in the source and target domains are very different in appearance, their corresponding output images may be very similar in spatial layout and localization.}.
\label{fig1}
\end{figure}

There are two kinds of approaches to solve such domain drift problem. One is the self-training-based approaches, which train the target data with pseudo-labels and the labeled source data achieve cross-domain alignment simultaneously. They improve the discriminative performance of the model on the target data. However, self-training methods usually need assign the pseudo-labels to the target data based on the confidence, and the obtained pseudo-labels are usually noisy. Such noise may weaken the generalization ability of the model. In addition, they require more pre-training steps, and hence the model complexity is generally high. On the other hand, adversarial learning-based UDA methods reduce domain differences by aligning the distribution of two domains in terms of appearances, features \cite{2019Synergistic,2020SSF}, or the output space  \cite{8578878,9157766,DBLP:conf/iccv/TsaiSSC19}. In 3S, we note that the output space usually contains rich information both globally and locally. That is, even if the images from two domains have different appearances, their segmentation output images will have substantial similarities in the overall spatial layout and locally (see, Fig\ref{fig1}). Based on this observation, Tsai et al. address the pixel-level domain adaption problem in the output space. However, the transfer mapping of the source and target domains obtained by network approximation may lead to overfitting phenomenon due to the large amount of network parameters, at the same time, it has few interpretability. 

Fortunately, OT, as a method of transferring information between samples of two domains, has been applied in the DA problem. It can be used to calculate the Wasserstein distance between two probability distributions to achieve distribution alignment. The feasibility of using OT for DA has been verrified in \cite{7586038}, where the discrete OT is used to match the edge distributions of two domains under class regularity constraints. Deep Joint Optimal Transfer (DeepJDOT) uses the same logic to transfer the source domain samples to the target domain through the coupling matrix \cite{2018arXiv180310081B}. Seg-JDOT \cite{10.3389/fncom.2020.00019} is an attempt to use this idea for medical image segmentation, which enables the source and the target domain samples sharing similar representations. Recently OT is applied to DA tasks using weighting methods \cite{9156476,9157821} to make the transmission of classification image samples more accurate. To our best knowledge, there is almost no OT-based method for the 3S, and hence in this paper, we introduce OT into 3S. 

In this paper, we use deep network to implement the UDA problem based on OT in SSS. Firstly, we input the source and target images into the segmentation network, which generates the output space via CNN due to its superiority of feature representation. Secondly, we utilize OT to achieve a more robust alignment of source and target domains in output space. By gradually adjusting OT coupling in segmentation network training, different weights (which can also be interpreted as an attention mechanism) are provided for sample information transmission between the source and target domain. The OT part achieves DA and interprets alignment process of source domain and target domain. At the same time, it reduces the number of network parameters. Moreover, low-level features and high-level features contain different feature information respectively, and transferring knowledge from the high-level to guide the adaptation of low-level features is necessary. Inspired by \cite{2014arXiv1409.5185L}, we perform DA on both low and high-level features in an integrated manner. It bridges the output of low-level features and high-level features. In other words, we construct a multi-scale segmentation network to perform adaptation in the output space at different feature levels.

The contributions of our work are as follows. 
\begin{itemize}
\item We utilize the OT to achieve the UDA and embed it into the semantic segmentation network, and further form an End-to-End CNN-based framework for 3S. It indeed reduces the number of network, as well as offers an explanation for the  semantic segmentation process.

\item we introduce a multiscale strategy to better model the low-level and high-level feature representations, which realizes the label transfer under different scales. 

\item We apply the model to the semantic segmentation of road scenes, and experimental results on real and synthetic datasets show that our proposed method has a better segmentation performance and a lower network complexity. 

\end{itemize}

This paper is organized as follows: Related work is introduced in Section 2.  Section 3 introduces some preliminaries of OT. Section 4 proposes our method and then presents a optimization strategy for solving this model. Section 5 presents experimental results implemented on synthetic and real data sets to verify the superiority of our proposed method over other
state-of-the-arts methods. Finally, Section 6 draws a conclusion.

\section{Related works}
\textbf{Deep learning based semantic segmentation.} Current semantic segmentation methods are basically based on deep learning methods, existing methods are divided into three categories of decoder-based, information fusion-based and recurrent neural network-based(RNN) approaches. In 2017, Shelhamer et al. proposed a semantic segmentation method based on fully convolutional neural networks (FCNs), which is the pioneering work in deep learning algorithms for semantic segmentation. It not only answers the question of how CNNs can be trained end-to-end for the semantic segmentation problem, but also effectively solves the problem of how to produce semantic predictions of pixel-level outputs for inputs of arbitrary size. However, the method also has some limitations, in that it tends to lose pixel location information when performing sampling, thus affecting segmentation accuracy. In 2017, the SegNet algorithm proposed by Badrinarayanan et al. designed an encoder-decoder network for semantic segmentation of road scenes, which improved image segmentation by retaining the pooling layer index when pooling was performed and reducing the training parameters through improvements, while being able to recover image boundary information more accurately, but its segmentation accuracy at object boundaries still needs further improvement.

To further optimize the semantic segmentation effect and make full use of the target spatial information, it is usually necessary to fuse different levels of information, which are mainly classified as pixel-level fusion, feature map fusion and multi-scale fusion. Compared with CNN, Conditional Random Fields (CRF) can better learn the correlation between pixels. Chen et al.\cite{2014Semantic} proposed DeepLabv1 to use the CRF model as an independent optimisation link in their network to achieve detail enhancement on segmentation results. DeepLabv2, proposed in 2016, introduced a pyramidal hole pooling (ASPP) module on top of DeepLabv1\cite{2016DeepLab} to select different sampling rates of convolution with holes to process feature maps and improve segmentation accuracy. DeepLabv3\cite{2017Rethinking} continued to optimize the ASPP structure and introduced the Resnet block module to effectively extract expressive features by cascading multiple cavity convolution structures. In 2018, Chen et al.\cite{DBLP:journals/corr/abs-1802-02611} proposed DeepLabv3+ which used DeepLabv3 as an encoder and the Xception network structure as a benchmark model, and designed a novel decoder structure that achieved better test results. Another common approach to information fusion for FCNs in semantic segmentation problems is to perform feature graph fusion. Feature graph fusion refers to combining the global feature graph extracted from the front layer of the network with the local feature graph extracted from the back layer. The main representative methods are ParseNet\cite{2015ParseNet}, SharpMask\cite{2016Learning}, PSPNet\cite{2016Pyramid}, etc. Another idea to achieve information fusion is the multi-scale fusion approach\cite{2015Multi}, \cite{2016A}, \cite{2014Predicting}, \cite{2016Multiscale}, where multiple networks at different scales are selected and combined with their predictions to produce a comprehensive output.

Recurrent neural networks (RNNs) combine pixel-level and local information for successful application in modelling global information and improving semantic segmentation results. Long short-term memory networks (LSTM)\cite{2016Long} and gated recurrent units (GRU)\cite{2016Investigating} are two mainstream RNN structures. In summary, the application of deep learning to solve the image semantic segmentation problem has gained rapid momentum. In addition to the above methods, many new ideas and methods have still emerged in recent years\cite{2017FoveaNet}, \cite{2018Learning}, \cite{2017Semi}, and have shown strong competitiveness.

\textbf{Domain adaptation based semantic segmentation.} The method proposed in this paper is based on CNN training to solve the UDA problem in semantic segmentation. UDA aims to reduce the domain difference between the labeled source domain and the unlabeled target domain, which could improve the generalization ability of the model on the target data. The current widely used UDA semantic segmentation methods are mainly divided into two categories: self-training to improve the adaptive ability of segmentation models using pseudo labeling and domain alignment through adversarial learning, and this paper mainly improves the UDA semantic segmentation method based on adversarial learning.

The semantic segmentation model of UDA based on adversarial learning mainly consists of two networks one network is used as a segmentation model to predict the output results, and the other network is used as a discriminator to determine whether the input is from the source segmentation output or the target segmentation output, and the training goal of the segmentation model is to deceive the discriminator so that the output space \cite{8578878,9157766,8954024,DBLP:conf/iccv/TsaiSSC19} or feature layer\cite{2019Synergistic,2020SSF} achieve domain alignment. 

In all these methods, the distribution of the source and target domains are aligned by the discriminator, a network structure. And in our work, the discriminator network is explained in terms of OT theory to achieve DA. In addition, the attention mechanism is used to obtain the target region that needs to be focused on, which can obtain more details and key information about the current task in semantic segmentation and improve the accuracy of segmentation results. Current researchers have proposed numerous models of attention mechanisms for semantic segmentation\cite{8100143,8953974,9385072}. The OT coupling computed in this paper can also be used as an attention mechanism, which is also an important improvement of the model.

\textbf{OT based domain adaptation.} Domain adaptation problems can be broadly classified into two categories, semi-supervised domain adaptation and unsupervised domain adaptation, according to whether the samples in the target domain are partially labeled. In the semantic segmentation problem, due to the high cost of obtaining pixel-level label annotations, the current research focuses on the unsupervised domain adaptation problem in semantic segmentation, and the key to the UDA problem is to solve the source and target domain alignment problem. 

The OT\cite{Ludger2009Optimal} was early proposed by the French mathematician Gaspard Monge to study the scheme of transforming one distribution to another at minimum cost, and OT provides efficient methods to compute the optimal mapping to transform one distribution to another, and to determine the distance between them. OT has been used for DA\cite{10.1007/978-3-662-44848-9_18,DBLP:conf/nips/PerrotCFH16} to learn transitions between domains, with associated theoretical guarantees\cite{2016Theoretical}. Deep Joint Optimal Transfer (DeepJDOT)\cite{2018arXiv180310081B} has performed well on many tasks by transferring the source domain samples to the target domain through a coupling matrix. Recently OT has been applied as a weighted method in domain adaptive tasks\cite{9156476,9157821} with more accurate transmission on image classification tasks. However, there is no method to make segmented image samples accurate transmission. In the SSS task, the computation of OT is a challenging problem, its computational cost being of order $O(n^3log(n))$, where $n$ is the number of samples. 

To solve this problem, \cite{2013Sinkhorn}introduced the entropy regular term and proposed the sinkhorn algorithm with $O(n^2)$ computational complexity in time and space, however, the computation of OT still requires a large cost when n is large. In this paper, we solve this problem by distributed training, thus achieving the exact transmission of segmented image samples between two domains.

\section{Preliminaries of OT}

In this section, we discuss the problem of achieving domain-adaptive optimal transmission. OT is a better method to establish nonlinear correspondence between samples in the source and target domains. In the first part, we introduced the OT optimization problem over discrete empirical distributions. Then, we discussed a regularized variant of this discrete optimal transport problem.

\subsection{Discrete OT}

First, we denote by $\mu_s$ and $\mu_t$ the discrete forms of the marginal probability distributions of samples $X_s$ and $X_t$ on the source and target domains, that is

\begin{equation}
\mu_s = \sum_{i=1}^{n_s}p^s_i\delta_{x^s_i},\quad  \mu_t=\sum_{i=1}^{n_t}p^t_i\delta_{x^t_i}
\end{equation}
where $\delta_{x_i}$ is the Dirac function at location $x_i\in\mathbb{R}^d$. $p^s_i$ and $p^t_i$ are probability masses accociated to the $i$th sample, they satisfy $\sum\nolimits_{i=1}^{n_s}p^s_i = \sum\nolimits_{i=1}^{n_t}p^t_i = 1$. The Kantorovitch problem is to find the probabilistic coupling $\gamma$ that minimizes the optimal transport distance and is defined in the following form:

\begin{equation}
\gamma^* = \mathop{\arg\min}\limits_{\gamma\in\prod(\mu_s,\mu_t)}\langle\gamma,C\rangle_F
\end{equation}
where $\langle.,.\rangle_F$ is the Frobenius dot product and $C$ is the cost matrix, and $C_{ij} = c(x^s_i, x^t_j)$ defines the cost of moving the probability mass $x^s_i$ to $x^t_j$. As previously detailed, this cost is choosen as the squared Euclidean distance between the two locations, i.e., $C_{ij}=\big\|x^s_i-x^t_j\big\|^2_2$. $\prod(\mu_s,\mu_t) = \{\gamma\in\mathbb{R}^{N_s\times N_t}_{+}\mid\gamma1=\mu_s,\gamma^T1=\mu_t\}$, which denotes the probability coupling space of two empirical distributions. A unique optimal solution $\gamma^*$ satisfying the above equation can be obtained by solving problem (2). Based on the above conditions, we have the Wasserstein distance \cite{Villani2014Optimal} on the probability space of the source and target domains as follows:

\begin{equation}
W(\mu_s,\mu_t) = \mathop{\min}\limits_{\gamma\in\prod(\mu_s,\mu_t)}\langle\gamma,C\rangle_F
\end{equation}

It should be pointed out that the Problem (2) is the original Kantorovitch problem, which is a constrained linear programming problem. However, it is complex and difficult to calculate, which can be solved by entropy regularization.

\begin{figure*}[t]
\centering
\includegraphics[width=1.0\textwidth]{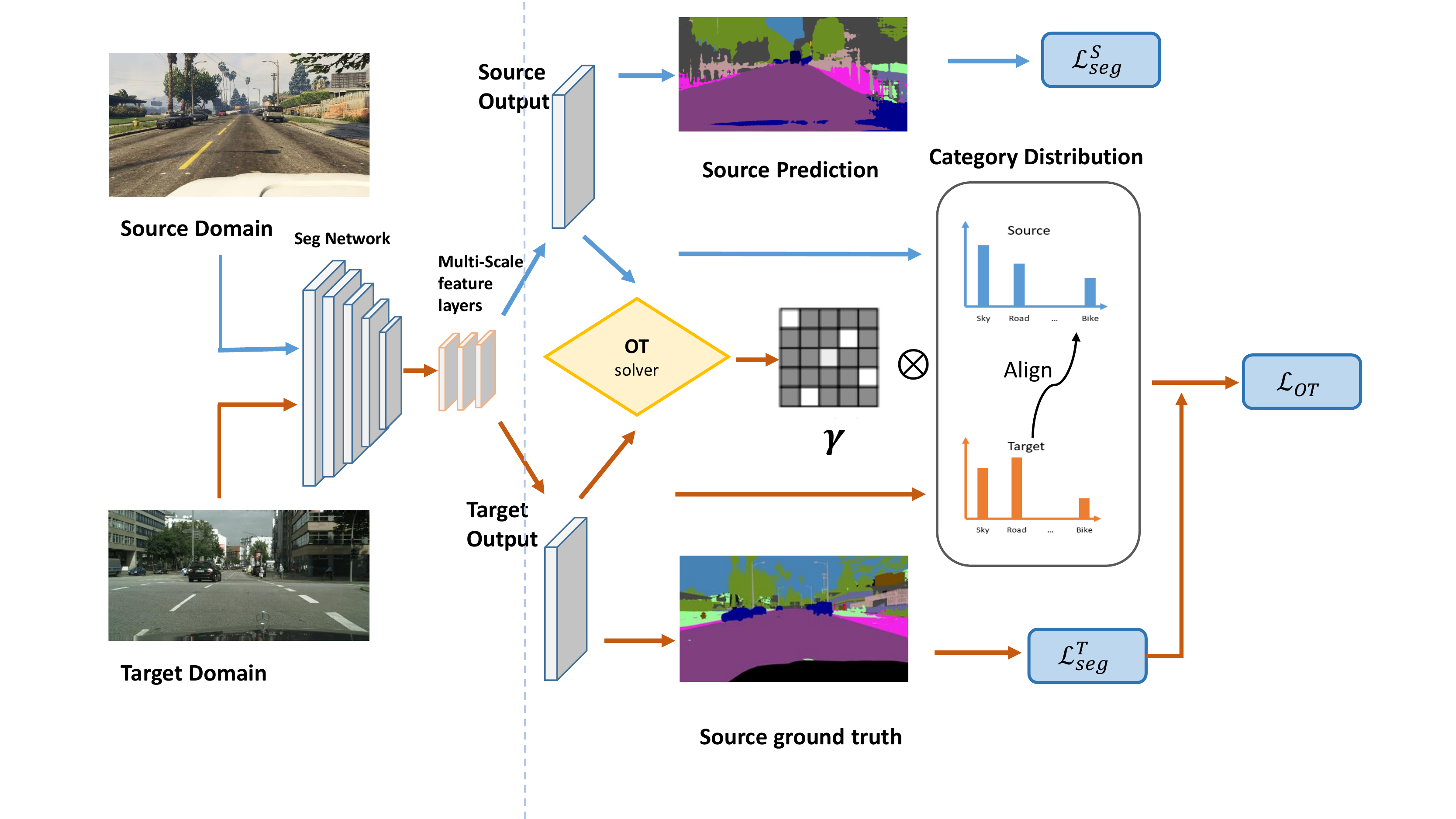} 
\caption{Overview of our proposed method. Given images from source and target domains, we pass them through the segmentation network to obtain output predictions. For source prediction, a segmentation loss is computed based on the source ground truth. To make target prediction closer to the source ones, We propose the OT module to achieve the distribution alignment of the source and target domains on the output space and to generate the OT loss. The source domain segmentation loss and OT loss are back propagated to the segmentation network.  }
\label{fig2}
\end{figure*}

\subsection{Regularized OT}
To solve the above computational complexity problem, Cuturi \cite{2013Sinkhorn} proposes to add entropy regularization term of probabilistic coupling to the equation. The entropy regularized version of the optimal transmission optimization problem is as follows:

\begin{equation}
\gamma^* = \mathop{\arg\min}\limits_{\gamma\in\prod(\mu_s,\mu_t)}\langle\gamma,C\rangle_F-\frac{1}{\lambda}\Omega(\gamma)
\end{equation}
where $\Omega(\gamma)=-\sum_{i,j}\gamma_{ij}$log$\gamma_{ij}$, since most elements in $\gamma^*$ have a high probability of being zero, the entropy regular term is added to the original OT problem in order to reduce its sparsity and make the transfer matrix smoother. $\lambda$ is the weight coefficient of the regular term. The larger $\lambda$ is, the lower the sparsity of $\gamma^*$. In other words, in the source domain tend to transmit their information to more target samples.

The entropy-regularized OT can be approximated by applying the Sinkhorn algorithm to find the approximate solution, which significantly reduces the computational complexity.

\section{Proposed Method}

In this section, we first introduce the proposed OT-based 3S framework and its basic framework, and then describe its process of implementing DA at the single-level and multi-level, respectively.

\subsection{The Framework of OT-based Scene Segmentation}
In this paper, we propose a UDA segmentation model based on OT. In the source domain, given the source images $X_S=\{x_s\in\mathbb{R}^{H\times W\times 3}\}^{n_s}_{j=1}$ and the corresponding pixel-wise one-hot labels $Y_S=\{y_s\in\{0,1\}^{H\times W\times C}\}^{n_s}_{j=1}$, while only target images $X_T=\{x_t\in\mathbb{R}^{H\times W\times 3}\}^{n_t}_{j=1}$ are available in the target domain. Here, $H$,$W$,$C$ denote the height, width of images and the number of classes, respectively. We aim to train a segmentation network that can correctly categorize pixels for target data $X_T$. 

As shown in Fig\ref{fig2}, we first input the source domain images $X_S$ with annotations into the segmentation network $G$ for training, and then use $G$ to make segmentation predictions for the images $X_T$ without annotations in the target domain to obtain the softmax output $P_t$. Since our goal is to make the segmentation predictions ($P_s$ and $P_t$) in the source and target domains closer, the feasibility of DA with OT has been verified\cite{7586038}. Here we use OT instead of the discriminator in the adversarial-based training network\cite{8578878} to achieve the alignment of the distributions of the two domains. That is, the segmentation network $G$ trained in the source domain is adapted to the target domain by encouraging the segmentation network $G$ to generate a segmentation distribution in the target domain that is similar to that in the source domain.

With the proposed method, we formulate the domain adaptation task containing two loss functions:

\begin{equation}
\mathcal{L}_{ST} = \mathcal{L}^S_{seg}(X_S,Y_S) +\mathcal{L}_{OT}(X_S, X_T, Y_S, Y_T)
\end{equation}
where $\mathcal{L}^S_{seg}(X_S,Y_S)$ is the cross-entropy loss using ground truth annotations in the source domain, and $\mathcal{L}_{OT}$ is the OT loss that adapts predicted segmentations of source images to the distribution of target predictions.

\subsection{Output Space Adaptation}

Compared with the high-dimensional feature space, the segmentation outputs are in the low-dimensional space, but they contain rich information such as the overall layout of the scene and local information. The basic idea of our proposed model is that regardless of whether the images are from the source or target domain, their segmentation should have strong spatial and local similarity. Tsai et al. \cite{8578878} exploits this property by an adversarial learning scheme to fit the low-dimensional softmax output of the segmentation prediction. In this paper, we further improve the adversarial learning scheme in \cite{8578878} to achieve the distribution alignment of segmentation outputs on the source and target domains by OT theory.

\subsubsection{Single-level Joint Distribution Optimal Transport}

Courty et al \cite{NIPS2017_0070d23b} proposed the Joint Distribution Optimal Transfer (JDOT) method for two-step adaptation (i.e., first adapting the representation and then learning the classifier on the adapted features) by directly learning the classifier embedded in the cost function $C$. The basic idea is to align joint feature/label distributions instead of only considering feature distributions. In the UDA semantic segmentation scenario, we use the weighted combination of costs in the output space and label space as a general loss for the distribution alignment of the source and target domains:

\begin{equation}
d(p^s_i, y^s_i;p^t_j,y^t_j) = \alpha c(p^s_i,p^t_j)+\beta \mathcal{L}(y^s_i,y^t_j)
\end{equation}

For the $i$-th source and $j$-th target sample, where $Y_S$ is the ground truth annotations for source images, $P_S = G(X_S)$ and $P_T= G(X_T)$ are the segmentation outputs of the source and target domains, respectively. In our problem, $c(p^s_i,p^t_j)$ is chosen as the Kullback-Leibler divergence between the class distributions of the segmentation output space of the source and target domains instead of $\ell^2_2$ distance, and $\mathcal{L}(y^s_i,y^t_j)$ is a cross-entropy classification loss. Parameters $\alpha$ and $\beta$ are two values weighing the contributions of distance terms. Since target labels $y^t_j$ are unknown, they are replaced by the segmentation output $p^t_j$. Based on the idea of OT, we want to match samples in the source and target domains that have similar segmentation outputs and share the same labels, and then we obtain the following optimization problem.
\begin{equation}
\mathcal{L}_{OT} =\mathop{\min}\limits_{G,\gamma\in\prod(\mu_s,\mu_t)}\langle\gamma,D_G\rangle_F
\end{equation}
where $D_G$ depends on $G$ and gathers all the pairwise cost $d(p^s_i, y^s_i;p^t_j,y^t_j)$, and $\prod(\mu_s,\mu_t)$ is the optimal transport scheme space from source domain to target domain. The solution to problem (7) can be achieved by minimizing the following objective function:
\begin{equation}
 \mathop{\min}\limits_{G,\gamma\in\prod(\mu_s,\mu_t)}\sum_i \sum_j \gamma_{ij}d(p^s_i, y^s_i,p^t_j)
\end{equation}
where $d(p^s_i, y^s_i;p^t_j,y^t_j) = \alpha KL(p^s_i \big{\|}p^t_j)+\beta\mathcal{L}(y^s_i,p^t_j) $, and $\mathcal{L}(y^s_i,p^t_j)$ is defined in the same way as the segmentation loss definition for the source domain below, and $\alpha$, $\beta$ are the parameters controlling the tradeoff between the two terms, as in equation(6). We use the KL divergence in the discrete case, $KL(p^s_i \big{\|}p^t_j)=\sum_{c}p^s_i log\frac{p^s_i}{p^t_j}$, that is, we want to use the segmentation output distribution $p^s_i$ of the source domain to guide the split output distribution $p^t_j$ of the target domain.

In addition we have to consider the segmentation performance of the segmentation network $G$ on the source domain samples, and according to equation (1), we define the segmentation loss in (5) as the cross-entropy loss for images from the source domain:
\begin{equation}
\mathcal{L}_{seg}(X_S,Y_S)=-\sum_{h,w}\sum_{c}Y^{(h,w,c)}_slog(P^{(h,w,c)}_s)
\end{equation}

\subsubsection{Multi-level Joint Distribution Optimal Transport}

Although performing OT alignment in the output space enables adaptive prediction in the target domain, the low-level feature and high-level feature spaces usually contain different information, and similar to the deep supervision approach using auxiliary loss for semantic segmentation in \cite{2014arXiv1409.5185L}, we added an OT module to the low-level feature space to enhance the adaption. The final training objective of the segmentation network G is:
\begin{small}
\begin{flalign*}
& \mathcal{L}_{ST}=\sum_{i}\lambda^i_{seg}\mathcal{L}^i_{seg}(X_S,Y_S)+\sum_i\lambda^i_{OT}\mathcal{L}^i_{OT}(X_S,X_T,Y_S) \quad (10)&
\end{flalign*}
\end{small}where $i$ denotes the network layer that predicts the segmentation output, and $\mathcal{L}^i_{seg}(X_S,Y_S)$ and $\mathcal{L}^i_{OT}(X_S,X_T,Y_S)$ remain in the same form as in (9) and (7), respectively. However, for large sample sizes the constraint of computing a full $\gamma$ yields a computationally problem, both in terms of memory and time complexity. In the next section, we propose a stochastic optimization method based on distributed training.

\subsection{Optimization Strategy}

In this part, we describe the approximate optimization procedure for solving problem (10). Equation (10) involves two variables to be optimized: the OT matrix $\gamma$ and the segmentation network $G$. Due to the huge sample size in the semantic segmentation scenario, computing a complete $\gamma$ generates computational problems in terms of memory and time complexity, we propose a small-batch distributed training method with random sampling from the source and target domains for optimization in each training.

There are two steps in each training optimization, first construct the cost matrix $C$ in equation (2), we construct the cost matrix $C$ by calculating the two-by-two similarity between the source and target domain samples in each small batch sampling. That is, $C_{ij}=\big\|x^s_i-x^t_j\big\|^2_2$, where $x^s_i$ and $x^t_j$ denote the initial input images of the source and target domains, respectively.

Then the OT coupling $\gamma$ is calculated by the sinkhorn algorithm, and when the $\gamma$ is fixed, the optimization of $G$ is a classical deep learning problem.

Based on the idea of OT theory, the calculated $\gamma_{ij}$ represents the probability of transmitting the information from the $i$-th sample in the source domain to the $j$-th sample in the target domain at the minimum transport cost. From equation (8), this probability value can also be regarded as the weight of aligning two samples in the loss function, which can be seen as an attention mechanism. That is, for the more similar samples in the source and target domains, a greater weight is assigned when performing the domain adaptive alignment, which further improves the efficiency of model adaption.

\begin{figure}[t]
\centering
\includegraphics[width=1.0\columnwidth]{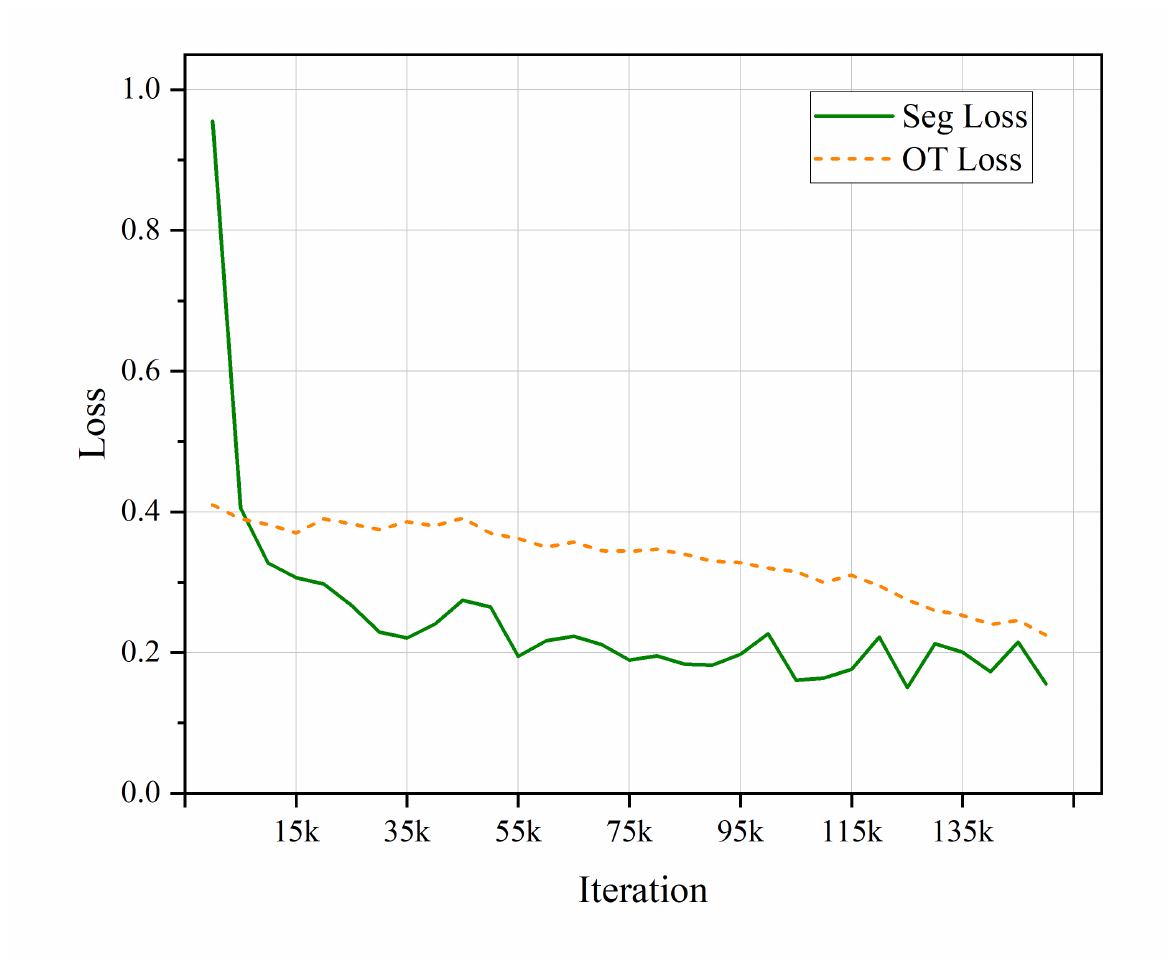} 
\caption{Loss variation diagram during model training.}
\label{fig3}
\end{figure}

We summarize this approach in Alogorithm 1, and the solution algorithm converges during the training process (see Figure\ref{fig3}).

\begin{algorithm}[!htbp]
\caption{Stochastic optimization algorithm}
\label{alg:algorithm}
\textbf{Input}: $x^s$:source domain samples, $x^t$:target domain samples, $y^s$:source domain labels\\
\textbf{Parameter}: $\alpha$, $\beta$, $\lambda^i_{seg}$, $\lambda^i_{OT}$\\
\textbf{Output}: $G$:segmentation network

\begin{algorithmic}[1] 
\FOR{each source batch($x^s_b$,$y^s_b$) and target batch($x^t_b$)}

\STATE Calculate $C$, and $C_{ij}=\big\|x^s_i-x^t_j\big\|^2_2$.
\STATE Calculate $\gamma$ for the given batch by Sinkhorn algorithm.
\STATE fix $\gamma$, and use gradient descent to update $G$.
\ENDFOR
\end{algorithmic}
\end{algorithm}

\begin{table*}[h]\centering
\begin{center}
\renewcommand\arraystretch{1.8}
\caption{Results of adapting GTA5 to CityScapes. We not only compare the results of performing OT alignment on the output space with other UDA semantic segmentation methods based on adversarial learning, but also compare the results of domain alignment without the inclusion of the OT module as an ablation experiment. Our baseline model is AdaptSegNet. Other previous state-of-the-art methods include CycleGAN\cite{2017Unpaired}, DLOW\cite{2019DLOW}, TGCF-DA+SE\cite{0Self}, SIBAN\cite{DBLP:conf/iccv/LuoLGYY19}, CLAN.}\label{table1}
\resizebox{18cm}{4.3cm}
{
\begin{tabular}{ccccccccccccccccccccc}
        \hline
        \multicolumn{21}{c}{GTA5$\to$CityScapes} \\
        \cline{1-21}
        Method &\rotatebox{90}{road} & \rotatebox{90}{sidewalk} & \rotatebox{90}{building} & \rotatebox{90}{wall} & \rotatebox{90}{fence} & \rotatebox{90}{pole} & \rotatebox{90}{light} & \rotatebox{90}{sign} & \rotatebox{90}{veg} & \rotatebox{90}{terrain} & \rotatebox{90}{sky} & \rotatebox{90}{person} & \rotatebox{90}{rider} & \rotatebox{90}{car} & \rotatebox{90}{truck} & \rotatebox{90}{bus} & \rotatebox{90}{train} & \rotatebox{90}{mbike} & \rotatebox{90}{bike} & mIoU \\ \hline
        Source only & 75.8 & 16.8 & 77.2 & 12.5 & 21.0 & 25.5 & 30.1 & 20.1 & 81.3 & 24.6 & 70.3 & 53.8 & 26.4 & 49.9 & 17.2 & 25.9 & 6.5 & 25.3 & 36.0 & 36.6 \\  \hline
        CycleGAN & 89.3 & 45.1 & 816 & 27.5 & 18.6 & 29.0 & 35.7 & 17.3 & 79.3 & 29.4 & 71.5 & 59.7 & 15.7 & 85.3 & 18.2 & 14.8 & 1.4 & 21.9 & 12.5 & 39.6 \\
        DLOW & 87.1 & 33.5 & 80.5 & 24.5 & 13.2 & 29.8 & 29.5 & 26.6 & 82.6 & 26.7 & 81.8 & 55.9 & 25.3 & 78.0 & 33.5 & 38.7 & 0.0 & 22.9 & 34.5 & 42.3 \\
        AdaptSegNet & 86.5 & 36.0 & 79.9 & 23.4 & 23.3 & 23.9 & 35.2 & 14.8 & 83.4 & 33.3 & 75.6 & 58.5 & 27.6 & 73.7 & 32.5 & 35.4 & 3.9 & 30.1 & 28.1 & 42.4 \\ 
        TGCF-DA+SE & 90.2 & 51.5 & 81.1 & 15.0 & 10.7 & 37.5 & 35.2 & 28.9 & 84.1 & 32.7 & 75.9 & 62.7 & 19.9 & 82.6 & 22.9 & 28.3 & 0.0 & 23.0 & 25.4 & 42.5 \\
        SIBAN & 88.5 & 35.4 & 79.5 & 26.3 & 24.3 & 28.5 & 32.5 & 18.3 & 81.2 & 40.0 & 76.5 & 58.1 & 25.8 & 82.6 & 30.3 & 34.4 & 3.4 & 21.6 & 21.5 & 42.6 \\
        CLAN & 87.0 & 27.1 & 79.6 & 27.3 & 23.3 & 28.3 & 35.5 & 24.2 & 83.6 & 27.4 & 74.2 & 58.6 & 28.0 & 76.2 & 33.1 & 36.7 & 6.7 & 31.9 & 31.4 & 43.2 \\ \hline
        Ours(only DA) & 81.2 & 24.8 & 78.6 & 19.9 & 22.0 & 26.7 & 32.6 & 26.5 & 78.1 & 14.5 & 74.3 & 59.7 & 29.1 & 73.2 & 28.5 & 22.0 & 3.4 & 26.0 & 32.0 & 39.6 \\
        Ours(single OT+DA) & 87.4 & 29.8 & 81.1 & 26.2 & 23.7 & 25.1 & 32.8 & 16.9 & 83.1 & 32.4 & 75.3 & 58.0 & 29.7 & 78.1 & 35.3 & 38.8 & 4.5 & 27.7 & 40.5 & 43.5  \\
        Ours(multi OT+DA) & 87.8 & 31.5 & 80.5 & 24.7 & 23.0 & 26.1 & 33.8 & 15.9 & 84.2 & 33.6 & 74.4 & 57.6 & 27.7 & 83.0 & 41.2 & 41.5 & 8.4 & 27.5 & 39.0 & {\bf44.3} \\ \hline
\end{tabular}
}

\end{center}
\end{table*}

\section{Experiments and Discussions}

The main purpose of our experiments is not only to evaluate the accuracy of our proposed method, but also to verify its lightness. In this section, we first introduce the datasets used in our experiments. Then we evaluate our method in different UDA semantic scene segmentation tasks and compare it with current methods, at the same time, we show our advantages in terms of model size. Finally, we visualize the scene segmentation results and the feature representations, which shows that our proposed method has better performance than compared semantic segmentation methods.

\subsection{Datasets}
Due to the high cost of obtaining real pixel-level labels of segmented images, we chose to use large-scale synthetic datasets with artificial annotations, GTA5\cite{2016Playing} and SYNTHIA\cite{7780721}. For model training, we used the real-world dataset Cityscapes\cite{7780719}as the target domain, hoping to adapt the trained model to it. Based on this setup, we conduct experimental comparision with three benchmark and four SOTA methods to validate our proposed approach.

We first evaluate the performance of our method on two synthetic-to-real semantic segmentation tasks, two synthetic datasets are GTA5\cite{2016Playing} and SYNTHIA\cite{7780721}, and the real dataset is CityScapes\cite{7780719}. And the two evaluation scenarios are GTA5$\to$CityScapes and SYNTHIA$\to$CityScapes. 

{\bf GTA5} dataset contains 24966 images captured from a video game based on the city of Los Angeles. Pixel-wise annotations with 33 classes are provided, but only 19 classes are utilized for compatibility with CityScapes. {\bf SYNTHIA} consists of 9400 synthetic images compatiable with the cityscapes annotated classes, and annotations with 13 classes are used for adaptation. {\bf CityScapes} is a real-world semantic segmentation dataset collected in driving scenarios. It contains 2975 unlabeled images in the training set, which are used as the target domain during training, and another 500 labeled images with manual annotations as the validation set for evaluation. 

In order to verify the performance of our proposed method, we conduct experimental comparison with state-of-the-art methods on the scene datasets, the methods used for comparison include AdaptSegNet, CycleGAN\cite{2017Unpaired}, DLOW\cite{2019DLOW}, TGCF-DA+SE\cite{0Self}, SIBAN\cite{DBLP:conf/iccv/LuoLGYY19}, CLAN, GIO-Ada\cite{2018Learning}, AdvEnt\cite{2019ADVENT}.

\begin{table}[h]\centering
\begin{center}
\renewcommand\arraystretch{1.4}
\caption{Performance gap between the adapted model and the fully-supervised(oracle)model. We first compare results with state-of-the-art methods using the VGG based model, and then show our result using the ResNet one. Our baseline model is AdaptSegNet\cite{8578878}. Other previous state-of-the-art methods include CDA\cite{2017Curriculum}, CyCADA\cite{pmlr-v80-hoffman18a}.}\label{table2}
\resizebox{8.6cm}{3.0cm}
{
\begin{tabular}{ccccc}
\hline
\multicolumn{5}{c}{GTA5$\to$CityScapes}  \\  \hline
Method & Baseline & Adapt & Oracle & mIoU Gap  \\ \hline 
CDA &  & 28.9 & 60.3 & -31.4 \\ 
CyCADA &  & 34.8 & 60.3 & -25.5 \\
AdaptSegNet(single) & {VGG-16} & 35.0 & 61.8 & -26.8 \\
Ours(single) &  & 35.8 & 62.4 & -26.6 \\ \hline
AdaptSegNet(multi) & {ResNet-101} &  42.4 & 65.1 & -22.7  \\
Ours(multi) &  &  44.3 & 66.2 & -21.9  \\ \hline
\end{tabular}
}

\end{center}
\end{table}

\begin{table*}
\begin{center}
\renewcommand\arraystretch{1.4}
\caption{Results of adapting SYNTHIA to CityScapes. We not only compare the results of performing OT alignment on the output space with other UDA semantic segmentation methods based on adversarial learning, but also compare the results of domain alignment without the inclusion of the OT module as an ablation experiment. Our baseline model is AdaptSegNet. Other previous state-of-the-art methods include CycleGAN\cite{2017Unpaired}, GIO-Ada\cite{2018Learning}, DLOW\cite{2019DLOW}, TGCF-DA+SE\cite{0Self}, SIBAN\cite{DBLP:conf/iccv/LuoLGYY19}, CLAN, AdvEnt\cite{2019ADVENT}.}\label{table3}
\resizebox{17cm}{4.5cm}
{
\begin{tabular}{ccccccccccccccccccccc}
        \hline
        \multicolumn{15}{c}{SYNTHIA$\to$CityScapes} \\
        \cline{1-15}
        Method &\rotatebox{90}{road} & \rotatebox{90}{sidewalk} & \rotatebox{90}{building} & \rotatebox{90}{light} & \rotatebox{90}{sign} & \rotatebox{90}{veg} & \rotatebox{90}{sky} & \rotatebox{90}{person} & \rotatebox{90}{rider} & \rotatebox{90}{car} & \rotatebox{90}{bus} & \rotatebox{90}{mbike} & \rotatebox{90}{bike} & mIoU \\ \hline
        Source only & 64.3 & 21.3 & 73.1 & 7.0 & 27.7 & 63.1 & 67.6 & 42.2 & 19.9 & 73.1 & 15.3 & 10.5 & 38.9 & 40.3 \\   \hline
        CycleGAN & 58.8 & 20.4 & 71.6 & 2.7 & 8.5 & 73.5 & 73.1 & 45.3 & 16.2 & 67.2 & 14.9 & 7.9 & 24.7 & 37.3 \\
        GIO-Ada & 78.3 & 29.2 & 76.9 & 10.8 & 17.2 & 81.7 & 81.9 & 45.8 & 15.4 & 68.0 & 15.9 & 7.5 & 30.4 & 43.0 \\ 
        SIBAN & 82.5 & 24.0 & 79.4 & 16.5 & 12.7 & 79.2 & 82.8 & 58.3 & 18.0 & 79.3 & 25.3 & 17.6 & 25.9 & 46.3 \\
        TGCF-DA+SE & 90.1 & 48.6 & 80.7 & 3.2 & 14.3 & 82.1 & 78.4 & 54.4 & 16.4 & 82.5 & 12.3 & 1.7 & 21.8 & 46.6 \\
        AdaptSegNet & 84.3 & 42.7 & 77.5 & 4.7 & 7.0 & 77.9 & 82.5 & 54.3 & 21.0 & 72.3 & 32.2 & 18.9 & 32.3 & 46.7 \\
        CLAN & 81.3 & 37.0 & 80.1 & 16.1 & 13.7 & 78.2 & 81.5 & 53.4 & 21.2 & 73.0 & 32.9 & 22.6 & 30.7 & 47.8 \\
        AdvEnt & 85.6 & 42.2 & 79.7 & 5.4 & 8.1 & 80.4 & 84.1 & 57.9 & 23.8 & 73.3 & 36.4 & 14.2 & 33.0 & 48.0 \\  \hline
        Ours(only DA) & 75.5 & 35.4 & 76.3 & 10.1 & 12.5 & 80.0 & 81.0 & 53.5 & 13.9 & 53.1 & 23.0 & 8.0 & 19.8 & 41.7 \\
        Ours(single OT+DA) & 84.2 & 18.0 & 77.4 & 28.9 & 17.4 & 80.9 & 72.2 & 52.9 & 24.1 & 69.8 & 24.4 & 21.0 & 34.6 & 46.6 \\
        Ours(multi OT+DA) & 87.6 & 43.8 & 80.6 & 11.2 & 12.1 & 81.1 & 81.2 & 56.7 & 20.1 & 74.8 & 33.7 & 16.8 & 34.2 & {\bf48.8} \\ \hline
\end{tabular}
}

\end{center}
\end{table*}

\begin{table}[h]\centering
\begin{center}
\renewcommand\arraystretch{1.3}
\caption{Performance gap between the adapted model and the fully-supervised(oracle)model. We first compare results with state-of-the-art methods using the VGG based model, and then show our result using the ResNet one. Our baseline model is AdaptSegNet\cite{8578878}. Other previous state-of-the-art methods include CDA\cite{2017Curriculum}, Cross-City\cite{2017No}.}\label{table4}
\resizebox{8.2cm}{2.5cm}
{
\begin{tabular}{ccccc}
\hline
\multicolumn{5}{c}{SYNTHIA$\to$CityScapes}  \\  \hline
Method & Baseline & Adapt & Oracle & mIoU Gap  \\ \hline 
CDA &  & 34.8 & 69.6 & -34.8 \\ 
Cross-City &  & 35.7 & 73.8 & -38.1 \\
AdaptSegNet(single) & {VGG-16} & 37.6 & 68.4 & -30.8 \\
Ours(single) &  & 38.5 & 69.0 & -30.5 \\ \hline
AdaptSegNet(multi) & {ResNet-101} &  46.7 & 71.7 & -25.0  \\
Ours(multi) &  &  48.8 & 72.5 & -23.7 \\ \hline
\end{tabular}
}

\end{center}
\end{table}

\begin{table}[h]\centering
\begin{center}
\renewcommand\arraystretch{1.4}
\caption{Comparison of the computational results of the space complexity (Params) and time complexity (Flops) of AdaptSegNet\cite{8578878} and our proposed model.
}\label{table5}

\resizebox{6.5cm}{1.35cm}
{
\begin{tabular}{ccc}
\hline

Method & Params & Flops  \\ \hline  
AdaptSegNet & 49.11M & 705468.37M \\
Ours & 43.55M & 651190.14M \\ \hline
\end{tabular}
}

\end{center}
\end{table}

\subsection{Network Architecture and Training}

{\bf Segmentation Network}. We adapt the DeepLab-v2\cite{2018DeepLab} framework with pre-trained ResNet-101\cite{7780459} encoder as our segmentation net. After the last layer, we use the Atrous Spatial Pyramid Pooling(ASPP)\cite{2018DeepLab} as the final classifier. Finally, we apply an up-sampling layer along with the softmax output to match the size of the input image.

We construct the above-mentioned segmentation net and apply the OT module to the output layer as our proposed model.

{\bf Network Training}. To train the proposed adaptation model, in each train batch, we first forward the source image $X_s$ to optimize the segmentation network for $\mathcal{L}^s_{seg}$ in (9) and generate the output $P_s$. For the target image $X_t$, we obtain the segmentation output $P_t$ and pass it along with $P_s$ for optimizing $\mathcal{L}_{OT}$ in (7). For the multi-level training objective in (10), we simply repeat the same procedure for each adaptation module.

Our method is implemented with the PyTorch library on Nvidia GPU GTX 2080Ti with 12GB memory. To train the segmentation network, we use the Stochastic Gradient Descent(SGD) optimizer with Nesterov acceleration where the momentum is 0.9 and the weight decay is 5 $\times$ $10^{-4}$. The initial learning rate is set as 2.5 $\times$ $10^{-4}$ and is decreased using the polynomial decay with power of 0.9 as mentioned in \cite{2018DeepLab}, and the maximum iteration number is 60000.

The performances of our method are evaluated by the widely utilized performance metrics, intersection-over-union(IoU) of each class and the mean IoU(mIoU).

\subsection{Results on GTA CityScapes}

We first evaluate the performance of our proposed method in the GTA5$\to$CityScapes scenario, and the corresponding results are listed in Table \ref{table1}. For a fair comparison, all the competed models adopt DeepLab-v2 network framework with pre-trained ResNet-101 as encoder. Our model is an improvement of the adversarial-based UDA method and shows better segmentation performance compared to other adversarial training-based domain alignment methods\cite{8578878,DBLP:conf/iccv/LuoLGYY19,2019DLOW,2017Unpaired,0Self}. Our model surpasses all these models with a promising mIoU of 44.3\%, it is demonstrated that our proposed method can effectively replace the discriminator part of the generative adversarial network and achieve better segmentation performance than it.

In addition, we use another factor to evaluate the adaptation performance, that is, measure how much gap is narrowed between the adaptation model and the fully-supervised model. So we train the model using annotated ground truths in the CityScapes datasets as the oracle results. The gap under different baseline models are showed in Table\ref{table2}. We compare the mIoU under two benchmark models, VGG-16 and ResNet-101, and it is clear that the gap is larger for the VGG one. This suggests that using a deeper benchmark model with larger capacity would be a greater improvement to our approach.

\subsection{Results on SYNTHIA CityScapes}

We then utilize SYNTHIA dataset as the source domain and display comparison results of our method and other state-of-the-art methods\cite{8578878,2018Learning,DBLP:conf/iccv/LuoLGYY19,2019ADVENT,2017Unpaired,0Self} that are adversarial training-based domain alignment on the validation set of CityScapes, as listed in the Table \ref{table3}. We consider the IoU and mIoU of a subset of 13 classes following the standard experimental setting\cite{2020Unsupervised}. Our method still achieves promising results in comparison to other competed methods. Specifically, the proposed method achieves 48.8\% mIoU of 13 categories.

In addition, with the same experimental setup as GTA5$\to$CityScapes, we measure how much gap is narrowed between the adaptation model and the fully-supervised model. So we train the model using annotated ground truths in the CityScapes datasets as the oracle results. The gap under different baseline models are showed in Table\ref{table4}. We compare the mIoU under two benchmark models, VGG-16 and ResNet-101, and it is clear that the gap is larger for the VGG one, too.

\subsection{Model size comparison}

To further demonstrate the simplicity of our proposed model, we compared the number of network parameters (Params) and the amount of computation (Flops), i.e., the spatial complexity and the temporal complexity of the model, between our method and the AdaptSegNet model\cite{8578878}. These two metrics for both models are presented in Table\ref{table5}. According to the computational results, our proposed model is further improved by reducing 11.32\% in spatial complexity and 7.7\% in temporal complexity compared to AdaptSegNet, while ensuring that the model segmentation results do not degrade and have some improvement.

\begin{figure*}[t]
\centering
\includegraphics[width=1.0\textwidth]{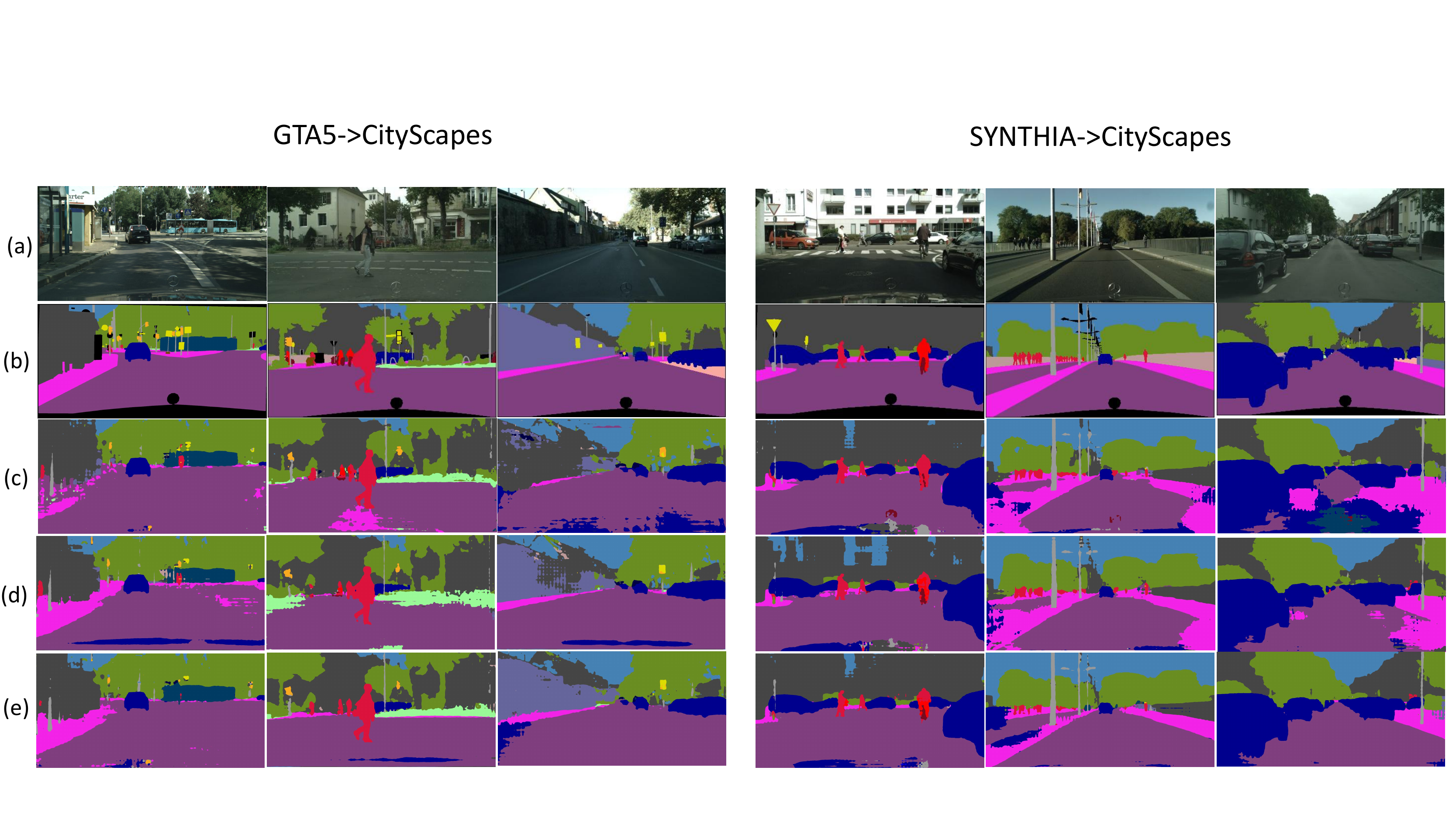} 
\caption{Qualitative results of UDA semantic segmentation. (a)Target image, (b)Ground truth, (c)source only model, (d)AdaptSegNet\cite{8578878}, (e)Ours. }
\label{fig4}
\end{figure*}

\subsection{Visualization Results}

{\bf Segmentation Visualization.} As illustrated in Figure\ref{fig4}, we provide some typical qualitative segmentation results of target data on two evaluation scenarios GTA$\to$CityScapes and SYNTHIA$\to$CityScapes. Obviously, the adversarial training-based UDA method\cite{8578878} could significantly promote the performance in comparison to the source model. Besides, our proposed model has better scalability to small-scale objectives(e.g.,'train'), and showed better segmentation performance on all other categories. Since OT is added as an attention mechanism in the adaptive process, the domain distribution alignment pays more attention to the category with higher similarity in both domains, so our proposed method provides more accurate supervision information and thus avoids some mislabeling and produces more reasonable segmentation results.

\begin{figure}[!htbp]
\centering
\includegraphics[width=1.0\columnwidth]{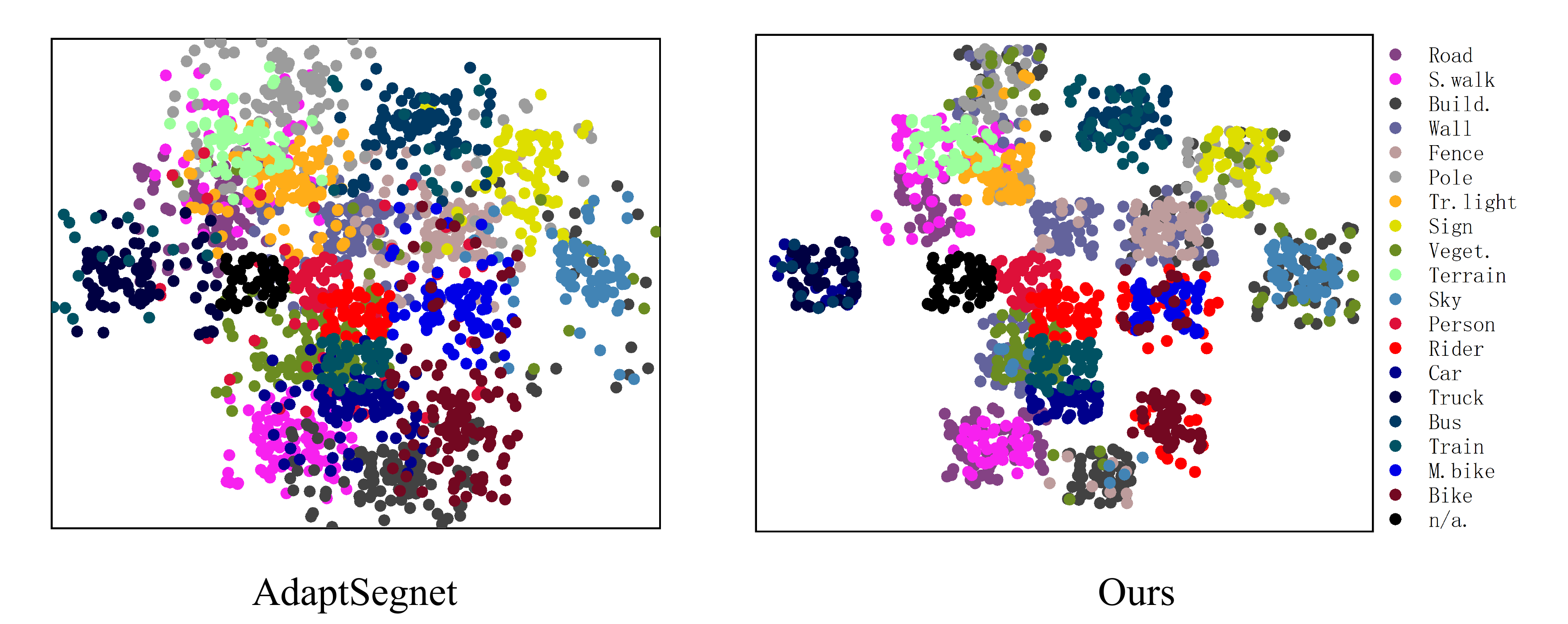} 
\caption{ The t-SNE visualization of embedded features.}
\label{fig5}
\end{figure}

{\bf Feature Visualization.} We use t-SNE\cite{2008Visualizing} to visualize the feature representations of our model and AdaptSegNet model\cite{8578878} as illustrated in Figure \ref{fig5}. It is observed that Our model learns a more compact and better separated pixel embedding, which suggests that our segmentation network can produce more discriminative features. This observation demonstrates that our method can provide correct supervision signal for target data through the OT part.

\section{Conclusion}

In this paper, we propose a UDA framework based on OT, and use a small size of labeled data to achieve semantic segmentation. Based on the fact that although the two images in the source and target domains have large differences in appearance, their segmentation outputs will have large similarities in the overall spatial layout or local structure, we achieve domain alignment of source and target domains on the segmented output. First we generate the output space via CNN, then we utilize OT to achieve DA of source and target domains in output space. In particular, the computed OT coupling matrix is used as an attention mechanism, which assigns higher weights to the samples with higher similarity in the two domains, and hence improves the adaptation of the model. Further, to better describe the multi-scale property of features, we construct multi-scale segmentation network on different feature layers to perform DA. Experimental results show that our proposed method performs favorably against three benchmark and four state-of-the-art methods, and visualization results under multiple domain adaptation 3S also show that our method has better performance.



\begin{thebibliography}{82}
\providecommand{\natexlab}[1]{#1}


\bibitem[{Ackaouy et~al.(2020)Ackaouy, Courty, Vallée, Commowick, Barillot,
  and Galassi}]{10.3389/fncom.2020.00019}
Ackaouy, A.; Courty, N.; Vallée, E.; Commowick, O.; Barillot, C.; and Galassi,
  F. 2020.
\newblock Unsupervised Domain Adaptation With Optimal Transport in Multi-Site
  Segmentation of Multiple Sclerosis Lesions From MRI Data.
\newblock \emph{Frontiers in Computational Neuroscience}, 14.


\bibitem[{{Bhushan Damodaran} et~al.(){Bhushan Damodaran}, {Kellenberger},
  {Flamary}, {Tuia}, and {Courty}}]{2018arXiv180310081B}
{Bhushan Damodaran}, B.; {Kellenberger}, B.; {Flamary}, R.; {Tuia}, D.; and
  {Courty}, N. 
\newblock {DeepJDOT: Deep Joint Distribution Optimal Transport for Unsupervised
  Domain Adaptation}.
\newblock arXiv:1803.10081.

\bibitem[{Bouville(2008)}]{c:22}
Bouville, M. 2008.
\newblock Crime and punishment in scientific research.
\newblock arXiv:0803.4058.

\bibitem[{Chen et~al.(2019)Chen, Dou, Chen, Qin, and Heng}]{2019Synergistic}
Chen, C.; Dou, Q.; Chen, H.; Qin, J.; and Heng, P.~A. 2019.
\newblock Synergistic Image and Feature Adaptation: Towards Cross-Modality
  Domain Adaptation for Medical Image Segmentation.
\newblock \emph{Proceedings of the AAAI Conference on Artificial Intelligence},
  33: 865--872.

\bibitem[{Chen et~al.(2018{\natexlab{a}})Chen, Zhu, Papandreou, Schroff, and
  Adam}]{DBLP:journals/corr/abs-1802-02611}
Chen, L.; Zhu, Y.; Papandreou, G.; Schroff, F.; and Adam, H.
  2018{\natexlab{a}}.
\newblock Encoder-Decoder with Atrous Separable Convolution for Semantic Image
  Segmentation.
\newblock \emph{CoRR}, abs/1802.02611.

\bibitem[{Chen et~al.(2014)Chen, Papandreou, Kokkinos, Murphy, and
  Yuille}]{2014Semantic}
Chen, L.~C.; Papandreou, G.; Kokkinos, I.; Murphy, K.; and Yuille, A.~L. 2014.
\newblock Semantic Image Segmentation with Deep Convolutional Nets and Fully
  Connected CRFs.
\newblock In \emph{International Conference on Learning Representations}.

\bibitem[{Chen et~al.(2016)Chen, Papandreou, Kokkinos, Murphy, and
  Yuille}]{2016DeepLab}
Chen, L.~C.; Papandreou, G.; Kokkinos, I.; Murphy, K.; and Yuille, A.~L. 2016.
\newblock DeepLab: Semantic Image Segmentation with Deep Convolutional Nets,
  Atrous Convolution, and Fully Connected CRFs.

\bibitem[{Chen et~al.(2018{\natexlab{b}})Chen, Papandreou, Kokkinos, Murphy,
  and Yuille}]{2018DeepLab}
Chen, L.~C.; Papandreou, G.; Kokkinos, I.; Murphy, K.; and Yuille, A.~L.
  2018{\natexlab{b}}.
\newblock DeepLab: Semantic Image Segmentation with Deep Convolutional Nets,
  Atrous Convolution, and Fully Connected CRFs.
\newblock \emph{IEEE Transactions on Pattern Analysis and Machine
  Intelligence}, 40(4): 834--848.

\bibitem[{Chen et~al.(2017{\natexlab{a}})Chen, Papandreou, Schroff, and
  Adam}]{2017Rethinking}
Chen, L.~C.; Papandreou, G.; Schroff, F.; and Adam, H. 2017{\natexlab{a}}.
\newblock Rethinking Atrous Convolution for Semantic Image Segmentation.

\bibitem[{Chen et~al.(2018{\natexlab{c}})Chen, Li, Chen, and
  Gool}]{2018Learning}
Chen, Y.; Li, W.; Chen, X.; and Gool, L.~V. 2018{\natexlab{c}}.
\newblock Learning Semantic Segmentation from Synthetic Data: A Geometrically
  Guided Input-Output Adaptation Approach.

\bibitem[{Chen et~al.(2017{\natexlab{b}})Chen, Chen, Chen, Tsai, and
  Min}]{2017No}
Chen, Y.~H.; Chen, W.~Y.; Chen, Y.~T.; Tsai, B.~C.; and Min, S.
  2017{\natexlab{b}}.
\newblock No More Discrimination: Cross City Adaptation of Road Scene
  Segmenters.
\newblock In \emph{2017 IEEE International Conference on Computer Vision
  (ICCV)}.

\bibitem[{Cheng, Dong, and Lapata(2016)}]{2016Long}
Cheng, J.; Dong, L.; and Lapata, M. 2016.
\newblock Long Short-Term Memory-Networks for Machine Reading.

\bibitem[{Choi, Kim, and Kim()}]{0Self}
Choi, J.; Kim, T.; and Kim, C. ????
\newblock Self-Ensembling with GAN-based Data Augmentation for Domain
  Adaptation in Semantic Segmentation.
\newblock \emph{International Conference on Computer Vision}.

\bibitem[{Clancey(1979)}]{c:79}
Clancey, W.~J. 1979.
\newblock \emph{{Transfer of Rule-Based Expertise through a Tutorial
  Dialogue}}.
\newblock {Ph.D.} diss., Dept.\ of Computer Science, Stanford Univ., Stanford,
  Calif.

\bibitem[{Clancey(1983)}]{c:83}
Clancey, W.~J. 1983.
\newblock {Communication, Simulation, and Intelligent Agents: Implications of
  Personal Intelligent Machines for Medical Education}.
\newblock In \emph{Proceedings of the Eighth International Joint Conference on
  Artificial Intelligence {(IJCAI-83)}}, 556--560. Menlo Park, Calif: {IJCAI
  Organization}.

\bibitem[{Clancey(1984)}]{c:84}
Clancey, W.~J. 1984.
\newblock {Classification Problem Solving}.
\newblock In \emph{Proceedings of the Fourth National Conference on Artificial
  Intelligence}, 45--54. Menlo Park, Calif.: AAAI Press.

\bibitem[{Clancey(2021)}]{c:21}
Clancey, W.~J. 2021.
\newblock {The Engineering of Qualitative Models}.
\newblock Forthcoming.

\bibitem[{Cordts et~al.(2016)Cordts, Omran, Ramos, Rehfeld, Enzweiler,
  Benenson, Franke, Roth, and Schiele}]{7780719}
Cordts, M.; Omran, M.; Ramos, S.; Rehfeld, T.; Enzweiler, M.; Benenson, R.;
  Franke, U.; Roth, S.; and Schiele, B. 2016.
\newblock The Cityscapes Dataset for Semantic Urban Scene Understanding.
\newblock In \emph{2016 IEEE Conference on Computer Vision and Pattern
  Recognition (CVPR)}, 3213--3223.

\bibitem[{Courty et~al.(2017{\natexlab{a}})Courty, Flamary, Habrard, and
  Rakotomamonjy}]{NIPS2017_0070d23b}
Courty, N.; Flamary, R.; Habrard, A.; and Rakotomamonjy, A. 2017{\natexlab{a}}.
\newblock Joint distribution optimal transportation for domain adaptation.
\newblock In Guyon, I.; Luxburg, U.~V.; Bengio, S.; Wallach, H.; Fergus, R.;
  Vishwanathan, S.; and Garnett, R., eds., \emph{Advances in Neural Information
  Processing Systems}, volume~30. Curran Associates, Inc.

\bibitem[{Courty, Flamary, and Tuia(2014)}]{10.1007/978-3-662-44848-9_18}
Courty, N.; Flamary, R.; and Tuia, D. 2014.
\newblock Domain Adaptation with Regularized Optimal Transport.
\newblock In Calders, T.; Esposito, F.; H{\"u}llermeier, E.; and Meo, R., eds.,
  \emph{Machine Learning and Knowledge Discovery in Databases}, 274--289.
  Berlin, Heidelberg: Springer Berlin Heidelberg.
\newblock ISBN 978-3-662-44848-9.

\bibitem[{Courty et~al.(2017{\natexlab{b}})Courty, Flamary, Tuia, and
  Rakotomamonjy}]{7586038}
Courty, N.; Flamary, R.; Tuia, D.; and Rakotomamonjy, A. 2017{\natexlab{b}}.
\newblock Optimal Transport for Domain Adaptation.
\newblock \emph{IEEE Transactions on Pattern Analysis and Machine
  Intelligence}, 39(9): 1853--1865.

\bibitem[{Cuturi(2013)}]{2013Sinkhorn}
Cuturi, M. 2013.
\newblock Sinkhorn Distances: Lightspeed Computation of Optimal Transportation
  Distances.
\newblock \emph{Advances in Neural Information Processing Systems}, 26:
  2292--2300.

\bibitem[{Du et~al.(2020)Du, Tan, Yang, Feng, Xue, Zheng, Ye, and
  Zhang}]{2020SSF}
Du, L.; Tan, J.; Yang, H.; Feng, J.; Xue, X.; Zheng, Q.; Ye, X.; and Zhang, X.
  2020.
\newblock SSF-DAN: Separated Semantic Feature Based Domain Adaptation Network
  for Semantic Segmentation.
\newblock In \emph{2019 IEEE/CVF International Conference on Computer Vision
  (ICCV)}.

\bibitem[{Eigen and Fergus(2014)}]{2014Predicting}
Eigen, D.; and Fergus, R. 2014.
\newblock Predicting Depth, Surface Normals and Semantic Labels with a Common
  Multi-Scale Convolutional Architecture.

\bibitem[{Engelmore and Morgan(1986)}]{em:86}
Engelmore, R.; and Morgan, A., eds. 1986.
\newblock \emph{Blackboard Systems}.
\newblock Reading, Mass.: Addison-Wesley.

\bibitem[{Fu et~al.(2019)Fu, Liu, Tian, Li, Bao, Fang, and Lu}]{8953974}
Fu, J.; Liu, J.; Tian, H.; Li, Y.; Bao, Y.; Fang, Z.; and Lu, H. 2019.
\newblock Dual Attention Network for Scene Segmentation.
\newblock In \emph{2019 IEEE/CVF Conference on Computer Vision and Pattern
  Recognition (CVPR)}, 3141--3149.

\bibitem[{Ganin et~al.(2016)Ganin, Ustinova, Ajakan, Germain, Larochelle,
  Laviolette, Marchand, and Lempitsky}]{2016Domain}
Ganin, Y.; Ustinova, E.; Ajakan, H.; Germain, P.; Larochelle, H.; Laviolette,
  F.; Marchand, M.; and Lempitsky, V. 2016.
\newblock Domain-Adversarial Training of Neural Networks.
\newblock \emph{Journal of Machine Learning Research}, 17(1): 2096--2030.

\bibitem[{Gong et~al.(2019)Gong, Li, Chen, and Gool}]{2019DLOW}
Gong, R.; Li, W.; Chen, Y.; and Gool, L.~V. 2019.
\newblock DLOW: Domain Flow for Adaptation and Generalization.
\newblock In \emph{2019 IEEE/CVF Conference on Computer Vision and Pattern
  Recognition (CVPR)}.

\bibitem[{Goodfellow et~al.(2014)Goodfellow, Pouget-Abadie, Mirza, Xu,
  Warde-Farley, Ozair, Courville, and Bengio}]{2014Generative}
Goodfellow, I.; Pouget-Abadie, J.; Mirza, M.; Xu, B.; Warde-Farley, D.; Ozair,
  S.; Courville, A.; and Bengio, Y. 2014.
\newblock Generative Adversarial Nets.
\newblock In \emph{Neural Information Processing Systems}.


\bibitem[{He et~al.(2016)He, Zhang, Ren, and Sun}]{7780459}
He, K.; Zhang, X.; Ren, S.; and Sun, J. 2016.
\newblock Deep Residual Learning for Image Recognition.
\newblock In \emph{2016 IEEE Conference on Computer Vision and Pattern
  Recognition (CVPR)}, 770--778.

\bibitem[{Hoffman et~al.(2018)Hoffman, Tzeng, Park, Zhu, Isola, Saenko, Efros,
  and Darrell}]{pmlr-v80-hoffman18a}
Hoffman, J.; Tzeng, E.; Park, T.; Zhu, J.-Y.; Isola, P.; Saenko, K.; Efros, A.;
  and Darrell, T. 2018.
\newblock {C}y{CADA}: Cycle-Consistent Adversarial Domain Adaptation.
\newblock In Dy, J.; and Krause, A., eds., \emph{Proceedings of the 35th
  International Conference on Machine Learning}, volume~80 of \emph{Proceedings
  of Machine Learning Research}, 1989--1998. PMLR.

\bibitem[{Kim and Byun(2020)}]{9157766}
Kim, M.; and Byun, H. 2020.
\newblock Learning Texture Invariant Representation for Domain Adaptation of
  Semantic Segmentation.
\newblock In \emph{2020 IEEE/CVF Conference on Computer Vision and Pattern
  Recognition (CVPR)}, 12972--12981.

\bibitem[{Laurens and Hinton(2008)}]{2008Visualizing}
Laurens, V. D.~M.; and Hinton, G. 2008.
\newblock Visualizing Data using t-SNE.
\newblock \emph{Journal of Machine Learning Research}, 9(2605): 2579--2605.

\bibitem[{Lecun et~al.(1998)Lecun, Bottou, Bengio, and Haffner}]{726791}
Lecun, Y.; Bottou, L.; Bengio, Y.; and Haffner, P. 1998.
\newblock Gradient-based learning applied to document recognition.
\newblock \emph{Proceedings of the IEEE}, 86(11): 2278--2324.

\bibitem[{{Lee} et~al.(2014){Lee}, {Xie}, {Gallagher}, {Zhang}, and
  {Tu}}]{2014arXiv1409.5185L}
{Lee}, C.-Y.; {Xie}, S.; {Gallagher}, P.; {Zhang}, Z.; and {Tu}, Z. 2014.
\newblock {Deeply-Supervised Nets}.
\newblock arXiv:1409.5185.

\bibitem[{Li et~al.(2020)Li, Zhai, Luo, Ge, and Ren}]{9157821}
Li, M.; Zhai, Y.-M.; Luo, Y.-W.; Ge, P.-F.; and Ren, C.-X. 2020.
\newblock Enhanced Transport Distance for Unsupervised Domain Adaptation.
\newblock In \emph{2020 IEEE/CVF Conference on Computer Vision and Pattern
  Recognition (CVPR)}, 13933--13941.

\bibitem[{Liu and Tuzel(2016)}]{2016Coupled}
Liu, M.~Y.; and Tuzel, O. 2016.
\newblock Coupled Generative Adversarial Networks.

\bibitem[{Liu, Rabinovich, and Berg(2015)}]{2015ParseNet}
Liu, W.; Rabinovich, A.; and Berg, A.~C. 2015.
\newblock ParseNet: Looking Wider to See Better.
\newblock \emph{Computer ence}.


\bibitem[{Luo et~al.(2019{\natexlab{a}})Luo, Liu, Guan, Yu, and
  Yang}]{DBLP:conf/iccv/LuoLGYY19}
Luo, Y.; Liu, P.; Guan, T.; Yu, J.; and Yang, Y. 2019{\natexlab{a}}.
\newblock Significance-Aware Information Bottleneck for Domain Adaptive
  Semantic Segmentation.
\newblock In \emph{2019 {IEEE/CVF} International Conference on Computer Vision,
  {ICCV} 2019, Seoul, Korea (South), October 27 - November 2, 2019},
  6777--6786. {IEEE}.



\bibitem[{Luo et~al.(2017)Luo, Zou, Hoffman, and Fei-Fei}]{2017Label}
Luo, Z.; Zou, Y.; Hoffman, J.; and Fei-Fei, L. 2017.
\newblock Label Efficient Learning of Transferable Representations across
  Domains and Tasks.

\bibitem[{{NASA}(2015)}]{c:23}
{NASA}. 2015.
\newblock Pluto: The 'Other' Red Planet.
\newblock \url{https://www.nasa.gov/nh/pluto-the-other-red-planet}.
\newblock Accessed: 2018-12-06.

\bibitem[{Niu et~al.(2022)Niu, Sun, Tian, Diao, Chen, and Fu}]{9385072}
Niu, R.; Sun, X.; Tian, Y.; Diao, W.; Chen, K.; and Fu, K. 2022.
\newblock Hybrid Multiple Attention Network for Semantic Segmentation in Aerial
  Images.
\newblock \emph{IEEE Transactions on Geoscience and Remote Sensing}, 60: 1--18.

\bibitem[{Pan et~al.(2020)Pan, Shin, Rameau, Lee, and Kweon}]{2020Unsupervised}
Pan, F.; Shin, I.; Rameau, F.; Lee, S.; and Kweon, I. 2020.
\newblock Unsupervised Intra-Domain Adaptation for Semantic Segmentation
  Through Self-Supervision.
\newblock \emph{Conference on Computer Vision and Pattern Recognition (CVPR)}.

\bibitem[{Pearlmutter(1995)}]{410363}
Pearlmutter, B. 1995.
\newblock Gradient calculations for dynamic recurrent neural networks: a
  survey.
\newblock \emph{IEEE Transactions on Neural Networks}, 6(5): 1212--1228.

\bibitem[{Perrot et~al.(2016)Perrot, Courty, Flamary, and
  Habrard}]{DBLP:conf/nips/PerrotCFH16}
Perrot, M.; Courty, N.; Flamary, R.; and Habrard, A. 2016.
\newblock Mapping Estimation for Discrete Optimal Transport.
\newblock In Lee, D.~D.; Sugiyama, M.; von Luxburg, U.; Guyon, I.; and Garnett,
  R., eds., \emph{Advances in Neural Information Processing Systems 29: Annual
  Conference on Neural Information Processing Systems 2016, December 5-10,
  2016, Barcelona, Spain}, 4197--4205.

\bibitem[{Pinheiro et~al.(2016)Pinheiro, Lin, Collobert, and
  Dollár}]{2016Learning}
Pinheiro, P.~O.; Lin, T.~Y.; Collobert, R.; and Dollár, P. 2016.
\newblock Learning to Refine Object Segments.

\bibitem[{Raj, Maturana, and Scherer(2015)}]{2015Multi}
Raj, A.; Maturana, D.; and Scherer, S. 2015.
\newblock Multi-Scale Convolutional Architecture for Semantic Segmentation.

\bibitem[{Redko, Habrard, and Sebban(2016)}]{2016Theoretical}
Redko, I.; Habrard, A.; and Sebban, M. 2016.
\newblock Theoretical Analysis of Domain Adaptation with Optimal Transport.

\bibitem[{Rice(1986)}]{r:86}
Rice, J. 1986.
\newblock {Poligon: A System for Parallel Problem Solving}.
\newblock Technical Report KSL-86-19, Dept.\ of Computer Science, Stanford
  Univ.

\bibitem[{Richter et~al.(2016)Richter, Vineet, Roth, and Koltun}]{2016Playing}
Richter, S.~R.; Vineet, V.; Roth, S.; and Koltun, V. 2016.
\newblock Playing for Data: Ground Truth from Computer Games.
\newblock \emph{Springer International Publishing}.

\bibitem[{Robinson(1980{\natexlab{a}})}]{r:80}
Robinson, A.~L. 1980{\natexlab{a}}.
\newblock New Ways to Make Microcircuits Smaller.
\newblock \emph{Science}, 208(4447): 1019--1022.

\bibitem[{Robinson(1980{\natexlab{b}})}]{r:80x}
Robinson, A.~L. 1980{\natexlab{b}}.
\newblock {New Ways to Make Microcircuits Smaller---Duplicate Entry}.
\newblock \emph{Science}, 208: 1019--1026.

\bibitem[{Ros et~al.(2016)Ros, Sellart, Materzynska, Vazquez, and
  Lopez}]{7780721}
Ros, G.; Sellart, L.; Materzynska, J.; Vazquez, D.; and Lopez, A.~M. 2016.
\newblock The SYNTHIA Dataset: A Large Collection of Synthetic Images for
  Semantic Segmentation of Urban Scenes.
\newblock In \emph{2016 IEEE Conference on Computer Vision and Pattern
  Recognition (CVPR)}, 3234--3243.

\bibitem[{Roy and Todorovic(2016)}]{2016A}
Roy, A.; and Todorovic, S. 2016.
\newblock A Multi-scale CNN for Affordance Segmentation in RGB Images.
\newblock In \emph{European Conference on Computer Vision}.

\bibitem[{Rüschendorf(2009)}]{Ludger2009Optimal}
Rüschendorf, L. 2009.
\newblock Optimal Transport. Old and New.
\newblock \emph{Jahresbericht der Deutschen Mathematiker-Vereinigung}, 111(2):
  18--21.

\bibitem[{Souly, Spampinato, and Shah(2017)}]{2017Semi}
Souly, N.; Spampinato, C.; and Shah, M. 2017.
\newblock Semi Supervised Semantic Segmentation Using Generative Adversarial
  Network.
\newblock In \emph{2017 IEEE International Conference on Computer Vision
  (ICCV)}.

\bibitem[{Sun and Saenko(2016)}]{2016Deep}
Sun, B.; and Saenko, K. 2016.
\newblock Deep CORAL: Correlation Alignment for Deep Domain Adaptation.
\newblock \emph{Springer International Publishing}.

\bibitem[{Tsai et~al.(2019)Tsai, Sohn, Schulter, and
  Chandraker}]{DBLP:conf/iccv/TsaiSSC19}
Tsai, Y.; Sohn, K.; Schulter, S.; and Chandraker, M. 2019.
\newblock Domain Adaptation for Structured Output via Discriminative Patch
  Representations.
\newblock In \emph{2019 {IEEE/CVF} International Conference on Computer Vision,
  {ICCV} 2019, Seoul, Korea (South), October 27 - November 2, 2019},
  1456--1465. {IEEE}.

\bibitem[{Tsai et~al.(2018)Tsai, Hung, Schulter, Sohn, Yang, and
  Chandraker}]{8578878}
Tsai, Y.-H.; Hung, W.-C.; Schulter, S.; Sohn, K.; Yang, M.-H.; and Chandraker,
  M. 2018.
\newblock Learning to Adapt Structured Output Space for Semantic Segmentation.
\newblock In \emph{2018 IEEE/CVF Conference on Computer Vision and Pattern
  Recognition}, 7472--7481.

\bibitem[{Villani and Cédric(2014)}]{Villani2014Optimal}
Villani; and Cédric. 2014.
\newblock \emph{Optimal transport : old and new}.
\newblock Optimal transport : old and new.



\bibitem[{Vu et~al.(2019)Vu, Jain, Bucher, Cord, and Perez}]{2019ADVENT}
Vu, T.~H.; Jain, H.; Bucher, M.; Cord, M.; and Perez, P. 2019.
\newblock ADVENT: Adversarial Entropy Minimization for Domain Adaptation in
  Semantic Segmentation.
\newblock In \emph{2019 IEEE/CVF Conference on Computer Vision and Pattern
  Recognition (CVPR)}.

\bibitem[{Wu and King(2016)}]{2016Investigating}
Wu, Z.; and King, S. 2016.
\newblock Investigating gated recurrent networks for speech synthesis.

\bibitem[{Xiao, Lim, and Ning(2016)}]{2016Multiscale}
Xiao, B.; Lim, S.~N.; and Ning, Z. 2016.
\newblock Multiscale fully convolutional network with application to industrial
  inspection.
\newblock In \emph{2016 IEEE Winter Conference on Applications of Computer
  Vision (WACV)}.

\bibitem[{Xin et~al.(2017)Xin, Jie, Wei, Liu, and Feng}]{2017FoveaNet}
Xin, L.; Jie, Z.; Wei, W.; Liu, C.; and Feng, J. 2017.
\newblock FoveaNet: Perspective-Aware Urban Scene Parsing.

\bibitem[{Xu et~al.(2020)Xu, Liu, Wang, Chen, and Wang}]{9156476}
Xu, R.; Liu, P.; Wang, L.; Chen, C.; and Wang, J. 2020.
\newblock Reliable Weighted Optimal Transport for Unsupervised Domain
  Adaptation.
\newblock In \emph{2020 IEEE/CVF Conference on Computer Vision and Pattern
  Recognition (CVPR)}, 4393--4402.

\bibitem[{Zhang, David, and Gong(2017)}]{2017Curriculum}
Zhang, Y.; David, P.; and Gong, B. 2017.
\newblock Curriculum Domain Adaptation for Semantic Segmentation of Urban
  Scenes.
\newblock In \emph{IEEE International Conference on Computer Vision}.

\bibitem[{Zhao et~al.(2016)Zhao, Shi, Qi, Wang, and Jia}]{2016Pyramid}
Zhao, H.; Shi, J.; Qi, X.; Wang, X.; and Jia, J. 2016.
\newblock Pyramid Scene Parsing Network.
\newblock In \emph{IEEE Computer Society}.

\bibitem[{Zhao et~al.(2017)Zhao, Shi, Qi, Wang, and Jia}]{8100143}
Zhao, H.; Shi, J.; Qi, X.; Wang, X.; and Jia, J. 2017.
\newblock Pyramid Scene Parsing Network.
\newblock In \emph{2017 IEEE Conference on Computer Vision and Pattern
  Recognition (CVPR)}, 6230--6239.

\bibitem[{Zhu et~al.(2017)Zhu, Park, Isola, and Efros}]{2017Unpaired}
Zhu, J.~Y.; Park, T.; Isola, P.; and Efros, A.~A. 2017.
\newblock Unpaired Image-to-Image Translation using Cycle-Consistent
  Adversarial Networks.
\newblock \emph{IEEE}.

\end{thebibliography}
\end{document}